\documentclass[10pt,twocolumn,letterpaper]{article}

\usepackage{cvpr}
\usepackage{times}
\usepackage{epsfig}
\usepackage{graphicx}
\usepackage{amsmath}
\usepackage{amssymb}

\usepackage[pagebackref=true,breaklinks=true,letterpaper=true,colorlinks,bookmarks=false]{hyperref}

 \cvprfinalcopy 

\def\cvprPaperID{****} 

\ifcvprfinal\fi
\begin{document}

\pagenumbering{gobble}

\title{Annotation Methodologies for Vision and Language Dataset Creation}

\author{Gitit Kehat\\
Computer Science Department \\
Brandeis University\\
Waltham, MA. 02453 USA\\
{\tt\small gititkeh@brandeis.edu}
\and
James Pustejovsky\\
Computer Science Department \\
Brandeis University\\
Waltham, MA. 02453 USA\\
{\tt\small jamesp@brandeis.edu}
}

\maketitle

\begin{abstract}
\vspace{-5mm}

   Annotated datasets are commonly used in the training and evaluation of tasks involving natural language and vision (image description generation, action recognition and visual question answering). However, many of the existing datasets reflect problems that emerge in the process of data selection and annotation. Here we point out some of the difficulties and problems one confronts when creating and validating annotated vision and language datasets. 
\end{abstract}

\vspace{-6mm}
\section{Introduction}
\vspace{-2mm}

 Recently, the use of Natural Language Processing (NLP) resources has become increasingly popular among the Computer Vision (CV) community, mostly thanks to the large-scale, easily accessible data from the web, and the growing popularity of online crowdsourcing  platforms, such as Amazon Mechanical Turk (AMT). Typical vision and language tasks that utilize such resources are image description generation, action and affordance recognition, and visual question answering (VQA).
 This expanding intersection has also been leading to the definition of new vision and language tasks, some of them, in their language-only form, have been well studied in the NLP community, such as VQA. In addition, some works aim to utilize both language and vision to create richer multi-modal semantic spaces and vectors \cite{kottur2015visual}.
 
Most   previous work analyzing vision and language datasets has dealt with the technical aspects of the collected datasets, rather than the data-gathering and annotation techniques used.\footnote{The survey in \cite{ferraro2015survey} presented a large listing of datasets, analyzed by number of images and structure and more advanced criterions like syntactic richness of the vocabulary and the density of the captions.}
Work about the annotation process of images mainly focused on the speed, efficiency and cost aspects of the process \cite{krishna2016embracing}. As far as we know, we are the first to discuss the issue of the quality of  \textit{annotation content} within vision and language datasets.
We present some of the major difficulties involving building and validating annotated vision and language resources, discuss their potential effects on results, and comment on combining NLP resources.


\begin{figure}
  \includegraphics[scale=0.25]{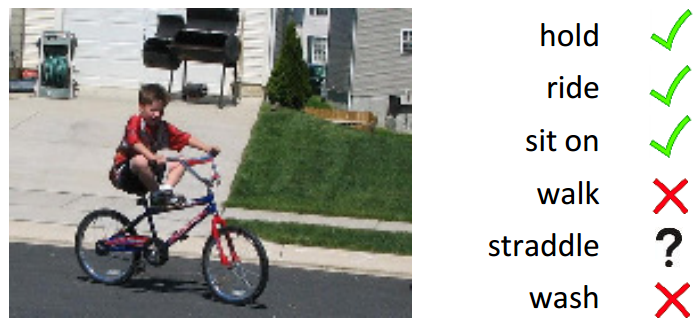}
 \includegraphics[scale=0.6]{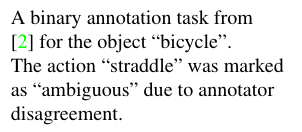}
\vspace{-4mm}
  \label{fig:HICO}
\end{figure}

\vspace{-9mm}

\section{Vision and language annotation tasks}
\vspace{-2mm}

Annotation tasks can have the form of multiple choice questions (closed) or open response ones, or a combination of the two.
The first case includes choosing ``yes"/``no" for a given option (e.g. a pair of an action and an object), choosing all true options from a given list (e.g. given an object, pick its attributes or actions) and more. The Second case includes supplying a full free-form sentence describing an image, picking free words to describe attributes or actions, or fill-in-the-blank a specific attribute or event \cite{yu2015visual}.
A combined case can allowed the user to add their own option to the list if needed \cite{ronchi2015describing}.
Tasks can ask the user to refer to the whole image or to only certain parts or objects, and sometimes require them to annotate the regions of some relevant objects. The images can be natural scenes taken by real people, or artificial scenes created with clip arts like in \cite{antol2015vqa}.
\subsection{Dimensions for Comparison}
\vspace{-2mm}

From the resources we surveyed, several dimensions emerged that can introduce potential weaknesses and inconsistencies during the creation of annotated datasets.


1. \textbf{Manual processing.} This concerns both  creating and validating the gathered data. Manual gold sets are widely used in the verification process. In \cite{Chao_2015_ICCV} for example, the set of verbs for each object is filtered manually, and verbs with similar meaning are grouped together manually. In \cite{gupta2015visual}, verbs for annotation are chosen and filtered manually, and annotations are examined manually too in order to penalize (what the authors see as)  common mistakes. Apart from being inefficient and unscalable, this also creates  author bias.
\newline 2. \textbf{Author bias.} This can occur when the gold set to evaluate annotation is created solely by the paper's authors. A way to minimize the bias is to start with a small set and bootstrap it according to an annotation guideline, as in \cite{gupta2015visual}, or to use as a gold set the majority of a subset of the annotations, as in \cite{gella2016unsupervised}.
\newline 3. \textbf{Limited or sparse vocabulary.} In \cite{Chao_2015_ICCV} and \cite{gupta2015visual} annotators can respond only to options created by a predefined set of verbs/actions, and have no way of introducing new terms into the dataset, imposing a heavier burden on initial annotation schema design. A possible solution can be to let the users add their own action terms, if necessary, as in  \cite{ronchi2015describing}.
Not limiting the users at all, on the other hand, can potentially create a dataset with very low and insignificant counts for each term. This can become a problem when the dataset is small, like in \cite{le2014tuhoi}. Grouping together similar terms is one way to solve it. However, such unification can potentially lead to a lose in the fine differences between textually described actions. An immediate preferable solution would be to largely extend the size of the dataset.
\newline 4. \textbf{Action/visual sense is not well defined.} This becomes a problem when dealing with action or affordance recognition. A well-defined annotation is necessary for both grounding the action to match external evaluations and resources, and for creating consistency among annotators of the same dataset.
This is especially crucial when dealing with data sets that were annotated with binary choices, such as \cite{Chao_2015_ICCV,gupta2015visual} (see Figure~\ref{fig:HICO}).
\newline 5. \textbf{Annotator's Attention.} When asked to describe an image, people tend to pick easy-to-describe relation (like ``man wearing a t-shirt") and start with the most salient parts of the image\cite{krishna2016visual}.
In \cite{le2014tuhoi} and \cite{ronchi2015describing} workers were asked to annotate action for specific already detected objects, forcing them to focus on objects otherwise might have been forgotten.
\newline 6. \textbf{Validation and Averaging.} These post-processing steps are especially important when the number of annotators per image is small. Most of the work surveyed validated the annotations to avoid  data that was corrupted for various reasons. In \cite{hodosh2013framing}, a short quiz verified the English level, and in \cite{gupta2015visual} two sets of test annotation questions were used to verify accuracy in submission time as well as  to filter out malicious turkers. 
Sometimes an averaging step is performed to simplify data from closed annotation tasks. However, when the averaging results in picking the majority of the annotators only (like in \cite{gupta2015visual}), or in a general non-informative ``ambiguous" tag (as in \cite{Chao_2015_ICCV}), or when non-agreement is completely ignored as in \cite{pirsiavash2014inferring}, this can lead to a possible information loss - just in the more interesting cases. A better solution would be to weigh according to several confidence levels chosen by the user for each annotation task (vs. just Y/N) (as in \cite{chao2015mining}), or according to the frequency (agreement) among annotators (as in \cite{ronchi2015describing}).

\vspace{-3mm}

\section{Using NLP Resources}

\vspace{-2mm}

Many of the more recent sources we reviewed used NLP tools and resources in some capacity. They are usually deployed as a ``single application" solution, in order to expand the vocabulary or enrich the vision-similarity equations with semantic data, without further use. However, it is important to understand the expected behavior of such tools.
For example, semantic vector space models (like Word2Vec) cannot distinguish between synonyms and antonyms, since they tend to appear in the same context \cite{santus2014taking}.
This is particularly important when dealing with binary attributes and yes/no questions.\footnote{The only work we found to have mentioned this issue was VQA \cite{antol2015vqa}.}
\vspace{-2mm}
\newline\newline
{\footnotesize \textbf{Acknowledgements:} This work was supported by Contract W911NF-15-C-0238 with the US Defense Advanced Research
Projects Agency (DARPA) and the Army Research Office (ARO).}

\vspace{-3mm}
{\footnotesize
\bibliographystyle{ieee}
\bibliography{egbib}
}
\end{document}